\documentclass{article}

\usepackage[final,main]{neurips_2026}

\usepackage[utf8]{inputenc}
\usepackage[T1]{fontenc}
\usepackage{amsmath}
\usepackage{amssymb}
\usepackage{amsfonts}
\usepackage{mathtools}
\usepackage{amsthm}
\usepackage{booktabs}
\usepackage{multirow}
\usepackage{graphicx}
\usepackage{enumitem}
\usepackage{float}
\usepackage{algorithm}
\usepackage{algorithmic}
\usepackage{placeins}
\usepackage{nicefrac}
\usepackage{microtype}
\usepackage{xcolor}
\usepackage{url}
\usepackage{hyperref}
\hypersetup{hidelinks}
\usepackage[capitalize,noabbrev]{cleveref}

\makeatletter
\renewcommand{\@notice}{}
\makeatother

\theoremstyle{plain}
\newtheorem{theorem}{Theorem}[section]
\newtheorem{proposition}[theorem]{Proposition}
\newtheorem{lemma}[theorem]{Lemma}
\newtheorem{remark}[theorem]{Remark}

\theoremstyle{definition}

\renewcommand{\vec}[1]{\boldsymbol{#1}}

\newcommand{\Z}{\mathbb{Z}}
\newcommand{\dist}{\mathrm{dist}}
\newcommand{\merge}{\mathsf{Merge}}
\newcommand{\aggr}{\mathsf{AGGR}}

\newcommand{\emb}{\mathsf{emb}}
\newcommand{\colorgnn}{\mathrm{ColorGNN}}
\newcommand{\colorlui}{\mathrm{ColorUID}}
\newcommand{\R}{\mathbb{R}}
\renewcommand{\Z}{\mathbb{Z}}
\newcommand{\D}{\mathcal{D}}
\renewcommand{\H}{\mathcal{H}}
\newcommand{\gen}{\mathrm{gen}}

\newcommand{\best}[1]{{\bfseries\boldmath #1}}

\title{Feature Augmentation of GNNs for ILPs: Local Uniqueness Suffices}

\author{
  Qingyu Han$^{1,2}$, Qian Li$^{2}$, Linxin Yang$^{3}$,
  Qian Chen$^{1,2}$, Qingjiang Shi$^{4}$, and Ruoyu Sun$^{2,3}$ \\
  $^{1}$School of Science and Engineering, The Chinese University of Hong Kong, Shenzhen, China \\
  $^{2}$Shenzhen International Center for Industrial and Applied Mathematics, Shenzhen Research \\
  Institute of Big Data, China \\
  $^{3}$School of Data Science, The Chinese University of Hong Kong, Shenzhen, China \\
  $^{4}$School of Software Engineering, Tongji University, Shanghai, China
}

\begin{document}

\maketitle

\begin{abstract}
  Integer Linear Programs (ILPs) are central to real-world optimizations but notoriously difficult to solve. Learning to Optimize (L2O) has emerged as a promising paradigm, with Graph Neural Networks (GNNs) serving as the standard backbone. However, standard anonymous GNNs are limited in expressiveness for ILPs, and the common enhancement of augmenting nodes with globally unique identifiers (UIDs) typically introduces spurious correlations that severely harm generalization. To address this tradeoff, we propose a parsimonious Local-UID scheme based on d-hop uniqueness coloring, which ensures identifiers are unique only within each node’s $d$-hop neighborhood. Building on this scheme, we introduce $\colorgnn$, which incorporates color information via color-conditioned embeddings, and $\colorlui$, a lightweight feature-level variant. We prove that for $d$-layer networks, Local-UIDs achieve the expressive power of Global-UIDs while offering stronger generalization. Extensive experiments show that our approach yields substantial and robust gains across ILP benchmarks. 
\end{abstract}

\section{Introduction}
Integer linear programs (ILPs) are important optimization problems with linear objectives and linear constraints, where variables are constrained to be integers. ILPs arise widely in engineering and operations research—including vehicle routing~\citep{ray2014multi,schobel2001model}, scheduling~\citep{ku2016mixed,meng2020mixed}, and facility location~\citep{holmberg1999exact,melkote2001capacitated}—yet remain computationally challenging. Traditional methods, most notably Branch and Bound \citep{boyd2007branch}, exhibit exponential worst-case complexity.

Learning to Optimize (L2O) has recently gained attention as a data-driven paradigm for addressing challenging optimization tasks. L2O methods learn optimization strategies
from existing problem instances to improve solution efficiency and quality~\citep{kruber2017learning,khalil2022mip,paulus2022learning,li2024pdhg}. For ILPs, Graph Neural Networks (GNNs) have emerged as a natural choice of architecture  \citep{gasse2019exact,nair2020solving,han2023gnn,huang2024contrastive,liu2025apollo}, since an ILP can be naturally represented as a bipartite graph: variable nodes on one side, constraint nodes on the other, with edges indicating variable participation in constraints.

However, standard GNNs have limited expressive power to solve ILPs \citep{Chen0WY23a,chen2025gnns}. A common way to overcome this limitation is to enrich the graph with additional distinguishing features, typically in the form of unique identifiers (UIDs) for nodes. Several feature augmentation strategies have been explored for ILPs, including random features \citep{Chen0WY23a} and positional embeddings \citep{han2023gnn,chen2024symilo}. These strategies can be viewed as instances of a Global-UID scheme, where each node is assigned a completely distinct identifier. In addition, to address variable symmetries, \cite{chen2025gnns} proposed an orbit-based augmentation method that assigns distinct UIDs to variables within the same symmetry orbit.

While feature augmentation can enhance expressiveness, it often comes at the cost of generalization, thereby requiring more training samples. Specifically, injecting extrinsic features or identifiers introduces irrelevant signals that increase the risk of overfitting: the model may rely on IDs as shortcut cues (see Appendix~\ref{app:identifier-shortcut}), memorizing training-specific patterns rather than learning structural regularities, thereby spuriously associating IDs with the target labels {\citep{keriven2021universality,bechler2024utilization,bechler2025towards}}. To alleviate this issue, augmentation should remain as parsimonious as possible. This motivates the following design principle:

\begin{center}
\textit{Feature augmentation should be parsimonious while retaining sufficient expressive power.}
\end{center}

\subsection{Our contribution}

In this paper, we introduce a simple yet effective feature augmentation scheme, termed \emph{Local-UIDs}, designed to be parsimonious without compromising expressive power. 
Our approach is motivated by two key observations:
\begin{itemize}[leftmargin=16pt]
\item Empirical GNNs for ILPs are typically \emph{shallow} (no more than four layers) \citep{gasse2019exact,han2023gnn, chen2024symilo,huang2024contrastive,chen2025gnns,liu2025apollo}, partly due to over-smoothing and over-squashing issues. 
\item In a $d$-layer GNN, the receptive field of each node is restricted to its $d$-hop neighborhood.
\end{itemize}
{Therefore, global uniqueness is stronger than necessary: for a $d$-layer GNN, it suffices to make identifiers unique only within each node's $d$-hop receptive field. This leads to our main principle:
\begin{center}
\emph{Local uniqueness within GNN receptive fields suffices for the expressive power of UIDs.}
\end{center}

Based on this principle, we make the following contributions.

\begin{itemize}[leftmargin=16pt]
\item \textbf{Local-UID scheme} We formalize the principle using $d$-hop unique coloring, which assigns colors so that each node can distinguish all vertices in its $d$-hop neighborhood. We show that this condition is equivalent to a proper coloring of the $2d$-th power graph, which leads
to a simple greedy construction. Compared with Global-UIDs, which use one distinct identifier per node, this scheme, termed \emph{Local-UIDs}, require only a number of colors controlled by the local $2d$-hop degree, and thus more parsimonious especially for sparse graphs and shallow GNNs. 

\item \textbf{Color-aware architectures} We instantiate Local-UIDs in two GNN architectures. The first, $\colorgnn$, injects color information via color-conditioned embedding mappings. This makes the model value-invariant: it automatically embeds colors without requiring predefined numeric encodings and can be seamlessly combined with other feature-augmentation techniques. The second, $\colorlui$, is a lightweight feature-level variant that treats the color as an additional node feature, thereby avoiding the potential parameter blow-up.

\item \textbf{Theoretical justification}.
We establish both expressivity and generalization guarantees, showing that Local-UIDs retain the expressive power of Global-UIDs, while significantly improving generalization and requiring far fewer training samples.
\item \textbf{Empirical validation}. We conduct extensive experiments demonstrating that Local-UIDs (i) achieve state-of-the-art performance on five ILP benchmarks, (ii) use substantially fewer identifiers than Global-UID baselines, and (iii) are robust across different GNN backbones, coloring choices, and coloring-injection schemes.
\end{itemize}
}

\section{Preliminaries}\label{sec:pre}
\paragraph{Basic Notations} Let $\R$ and $\Z$ denote the sets of real and integer numbers respectively. We use bold lowercase letters (i.e., $\vec{x}$) to denote vectors, bold uppercase letters (i.e., $\vec{A}$) to denote matrices. 
 Let $G=(V,E)$ be a graph. Given a node $v\in V$, we define its \emph{{$d$-hop neighborhood}} as
\[
{N_d(v):=\{u\in V\setminus\{v\} \mid \dist(u,v)\leq d\}}, 
\]
where $\dist(u,v)$ is the shortest-path distance. 
We let {$B_d(v)$} denote the induced subgraph on {$N_d(v)\cup\{v\}$}.
For a vertex $v\in V$, we denote its degree as $\deg(v) = |N_1(v)|$, and more generally its $d$-hop degree as $\deg_d(v)=|N_d(v)|$. Let $\Delta_d:=\max_v \deg_d(v)$. The $d$-th power of $G$, denoted $G^d=(V,E^d)$, is defined as $(u,v)\in E^d$ if and only if $\mathrm{dist}_G(u,v)\leq d$.

\paragraph{Integer Linear Programming} An integer linear program (ILP) takes the following form:
\[
\min_{\vec{x}} \vec{c}^T\vec{x},\quad\quad\text{s.t. } \vec{A}\vec{x}\leq \vec{b},\quad \vec{x}\in \Z^n   
\]
where $\vec{x}\in \Z^n$ are integer variables, $\vec{A}\in\R^{m\times n}$ is the constraint matrix, $\vec{b}\in \R^{m}$ and $\vec{c}\in \R^{n}$ are coefficients. An ILP instance $(\vec{A},\vec{b},\vec{c})$ can be naturally encoded as a labeled bipartite graph $G=(V\cup U,E,L,W)$: 
\begin{itemize}[leftmargin=16pt]
\item A left node $v_i$ represents a variable $x_i$, a right node $u_j$ represents a constraint $\vec{A}_{j,:}\vec{x}\leq b_j$, and they are connected if the variable participates in the constraint; 
\item The label of a left node $v_i$ is $L(v_i)=c_i$, and the label of a right node $u_j$ is $L(u_j)=b_j$;
\item The weight of an edge $e=(v_i,u_j)$ is $W(e)=A_{ij}$.
\end{itemize}

\paragraph{Graph Neural Network} A $d$-depth Graph Neural Network (GNN) can be viewed as a parameterized function as follows: it takes a labeled graph $G=(V,E,L,W)$ as input,
\begin{enumerate}[leftmargin=16pt]
\item $\vec{h}^{(0)}_v:=\emb({\vec{L}_v})$, where $\emb(\cdot)$ is a learnable linear mapping.
\item For $k=1$ to $d$, compute 
\begin{align}
\label{def_mvk}
\vec{h}_{v}^{(k)}:=\merge^{(k)}\left(\vec{h}_{v}^{(k-1)},\aggr({\{(\vec{h}^{(k-1)}_{u},W(u,v))\mid u\in N_1(v)\}}) \right),
\end{align}
where $\merge^{(k)}$ is a multi-layer perceptron (MLP), and $\aggr$ is a fixed function which is permutation-invariant with respect to its arguments.
\item Finally, each node $v$ outputs $\vec{h}^{(d)}_v$.
\end{enumerate}

\section{Methodology}
In this section, we present our Local-UID scheme, which is based on $d$-hop unique coloring, and introduce a novel color-aware architecture $\colorgnn$ and its lightweight variant $\colorlui$. All proofs are provided in the Appendix~\ref{all_proof}.
\subsection{Local-UID Scheme}\label{subsec:local-uid}

Local-UID scheme is based on the notion of \emph{$d$-hop unique coloring} of a graph $G=(V,E)$, defined as a mapping \(C\colon V\rightarrow\mathbb{N}\) such that: for any $v\in V$, the vertices in $N_d(v)\cup\{v\}$ have distinct colors. It turns out that this notion can be exactly characterized by a $(2d)$-hop coloring. Recall that a $d$-hop coloring is a map \(C\) such that $C(u)\neq C(v)$ for any two distinct \(u,v\in V\) with \(\operatorname{dist}(u,v)\le d\). A $d$-hop coloring can equivalently defined as a proper coloring of the power graph \(G^d\).

\begin{proposition}\label{prop:dhop1}
For any graph $G$, a coloring is $d$-hop unique if and only if it is a $(2d)$-hop coloring (equivalently, a proper coloring of $G^{2d}$).
\end{proposition}

Based on Proposition \ref{prop:dhop1}, we construct a $d$-hop unique coloring by performing a greedy algorithm to find a proper coloring of $G^{2d}$, depicted in Algorithm~\ref{alg:dhop-coloring}. Here, $\bot$ denotes the uncolored state, and BFS denotes Breadth-First Search. We sort the vertices of $G^{2d}$ in non-increasing degree order and apply greedy coloring accordingly, which empirically reduces the number of colors required. We have the following lemma.

\begin{lemma}\label{lem:dhop2}
For any graph $G$, Algorithm~\ref{alg:dhop-coloring} constructs a $d$-hop unique coloring using at most $\Delta_{2d}+1$ colors in $\tilde{\Theta}(|G^{2d}|)=\tilde{O}(|V|\cdot \Delta_{2d}){\leq\tilde{O}(|V|^2)}$ time.
\end{lemma}

Lemma~\ref{lem:dhop2} implies that our Local-UID scheme can substantially reduce the number of identifiers: in contrast to Global-UIDs assigning $|V|$ distinct identifiers, it requires only at most $\Delta_{2d}+1$ distinct colors. The savings are particularly significant when $d$ is small and the graph is sparse. In Section \ref{sec:methodab} (see Table \ref{tab:ab2}) , we explicitly compare the number of identifiers required on ILP benchmarks.
{\begin{remark}[Choosing the coloring radius in practice]\label{remark:radius}
In practice, the coloring radius $r$ can be chosen smaller than $2d$. Our expressivity result (Theorem \ref{thm:d-hop}) shows that choosing $r=2d$ is sufficient to recover the expressive power of Global-UIDs. This should be interpreted as a worst-case sufficiency guarantee, not a requirement for good empirical performance.
In fact, a smaller radius $r<2d$ may already provide good distinguishability for a given task (see Remark \ref{remark:1-hop}), while using fewer colors, reducing the parameter cost of $\colorgnn$, and lowering the preprocessing overhead. We therefore regard $r$ as a practical hyperparameter controlling the tradeoff between local distinguishability and parsimony.
\end{remark}
}
\begin{algorithm}[t]
  \caption{Greedy $d$-hop unique coloring}
  \label{alg:dhop-coloring}
  \textbf{Input}: A graph $G=(V,E)$, and an integer $d\geq 1$\\
  \textbf{Output:} Coloring $C\!: V \to \mathbb{N}$
  \begin{algorithmic}[1]
    \FOR{$v \in V$}
    \STATE {$N_{2d}(v) \leftarrow \mathrm{BFS}(G, v, 2d)$}
    \ENDFOR
    \STATE\text{construct}  $G^{2d}$ with $\{N_{2d}(v)\}$
   \STATE $\tilde{V}\leftarrow$ vertices of $V$ sorted in nonincreasing order of $\deg_{2d}(v)$
    \STATE $C(v) \leftarrow \bot$\textbf{ for all} $v \in V$
    \STATE $\text{Forbidden} \leftarrow \emptyset$
    \FOR{$v \in \tilde{V}$}
      \STATE $\text{Forbidden} \leftarrow \{\, C(u) \mid u \in N_{2d}(v) \land C(u) \neq \bot \,\}$
      \STATE $c \leftarrow 0$
      \WHILE{$c \in \text{Forbidden}$}
        \STATE $c \leftarrow c + 1$
      \ENDWHILE
      \STATE $C(v) \leftarrow c$
    \ENDFOR
    \STATE \textbf{return} $\{C(v)|v\in V\}$
  \end{algorithmic}
\end{algorithm}

\subsection{Color-Aware Architectural Design of GNNs}
Given a $d$-hop unique coloring produced by Algorithm~\ref{alg:dhop-coloring}, we introduce a novel GNN architecture, termed $\colorgnn$, that integrates the coloring into the network design.

\paragraph{ ColorGNN Architecture}  The key idea is to incorporate colors via color-conditioned embedding mappings, thereby avoiding the need for predefined numeric encodings. Moreover, $\colorgnn$ remains fully compatible with other feature-augmentation techniques. Formally, given a labeled graph $G=(V,E,L,W)$ and a coloring $C:V\rightarrow\mathbb{N}$, $\colorgnn$ proceeds as follows:
\begin{enumerate}[leftmargin=16pt]
\item $\vec{h}^{(0)}_v:=\emb_{C(v)}(\vec{L}_v)$, where $\emb_{C(v)}(\cdot)$ is a learnable linear transformation specific to color $C(v)$.
\item For $k=1$ to $d$, compute 
{\[
\vec{h}_{v}^{(k)}:=\merge^{(k)}\left(\vec{h}_{v}^{(k-1)},\aggr({\{(\vec{h}^{(k-1)}_{u},W(u,v))\mid u\in N_1(v)\}})\right)
\]}

\item Finally, each node $v$ outputs $\vec{h}^{(d)}_v$.
\end{enumerate}

\textit{Design choice}: We condition on the embedding functions rather than the aggregation $\aggr$ or merge functions for two reasons. First, $\aggr$ is typically required to remain permutation-invariant; conditioning it on colors risks violating this property or entangling UIDs with neighborhood multisets, which can hurt transferability across architectures. {Nevertheless, we include this variant in our experiments and report its performance for completeness.} Second, conditioning on $\merge$ offers no additional expressive power beyond embedding-level conditioning. In practice, color-conditioning consistently improves performance; the $\emb$-based $\colorgnn$ as well as the $\merge$-based and {$\aggr$-based variants} all achieve better results (see Table \ref{tab:ab1}).

\paragraph{ColorUID Architecture} {In $\colorgnn$, colors are used to select color-specific embedding maps: 
nodes with color $c$ share the share map $\emb_c$. Since $\colorgnn$ assigns a separate embedding map to each color, }

the parameter count scales with the number of colors $|C|$; for graphs with large $|C|$, this may cause parameter blow-up. To address this, we propose a lightweight variant, termed $\colorlui$. Architecturally, 
{$\colorlui$ treats the color $C(v)$ as an additional input feature, namely $\vec{h}^{(0)}_v:=\emb([\vec{L}_v; C(v)])$}, 

using a numeric representation consistent with prior UID schemes {~\citep{chen2022representing,han2023gnn,chen2025gnns}}. 
Empirical experiments show that $\colorlui$ consistently improves accuracy across all benchmark datasets, demonstrating the effectiveness { and robustness} of the Local-UID framework.

\section{Theoretical Justification}\label{sec:theory}
\subsection{Expressiveness Analysis}
We show that a $d$-hop unique coloring provides the same expressive power as global unique identifiers (Theorem~\ref{thm:d-hop}). Our proof leverages the equivalence between $d$-layer GNNs and $d$-round distributed algorithms, see e.g., ~\citep{loukas2020graph,sato2019approximation}. 
We extend this equivalence to the setting of colored GNNs in Lemma~\ref{prop:gnn=lui}. Specifically, in the standard distributed computing model, the network is represented by a graph $G=(V,E)$, where (i) each node corresponds to a processor and (ii) each edge $(u,v)\in E$ corresponds to a bidirectional communication channel. The computation proceeds in synchronous rounds: in each round, all processors execute a message-passing procedure to exchange information with their neighbors, followed by an identical local computation. Unless otherwise stated, we consider \emph{anonymous} distributed algorithms, where nodes do not have identifiers. 
\begin{lemma}
\label{prop:gnn=lui}
A $d$-layer $\colorgnn$, a $d$-layer $\colorlui$, and a $d$-round distributed algorithm equipped with the same coloring are equivalent in expressive power.
\end{lemma}

\begin{theorem}[Local uniqueness suffices]
\label{thm:d-hop}
If a function $f(G)$ can be computed by a $d$-layer GNN with global unique identifiers, then it can also be computed by a $d$-layer $\colorgnn$ or $\colorlui$ with a $2d$-hop coloring, or equivalently a $d$-hop unique coloring.
\end{theorem}

\begin{remark}\label{remark:1-hop}
{Even 1-hop uniquely colored $\colorgnn$ and $\colorlui$ can solve several nontrivial graph problems expressible as ILPs, including maximal independent set and maximal matching. Specifically, 
\citet{emek2014anonymous} showed that every problem solvable by a randomized anonymous Las Vegas distributed algorithm can also be solved by a deterministic anonymous distributed algorithm provided with a 2-hop coloring. By Proposition \ref{prop:dhop1} and Lemma \ref{prop:gnn=lui}, 

it follows that any such problem can be solved by 1-hop uniquely colored $\colorgnn$ and $\colorlui$. This applies in particular to maximal independent set \citep{alon1986fast,luby1985simple} and maximal matching \citep{emek2014anonymous}, which are known to admit such randomized anonymous algorithms.} 

\end{remark}

\subsection{\texorpdfstring{{Generalization Analysis}}{Generalization Analysis}}\label{sec:generalization}
{We analyze the generalization behavior of $\colorgnn$ for both node-level and graph-level prediction. Node-level prediction corresponds to the main learning task considered in this paper, namely predicting the solution value associated with each variable node in the bipartite ILP graph. We further include graph-level prediction to cover instance-level tasks, such as feasibility prediction~\citep{Chen0WY23a}. Our empirical evaluation focuses on node-level solution prediction.}

We consider the standard supervised learning setting with $N$ i.i.d. labeled graph samples drawn from an unknown distribution $\D$. The architecture, including depth, width, and aggregator, is fixed. 
{The coloring is generated by a fixed deterministic rule $\mathcal{C}$: given the same input graph topology, the rule always returns the same coloring, using a common palette of size $|\mathcal{C}|$. Our bounds
therefore isolate the effect of the color-palette size.}

We assume that all real-valued parameter entries are stored with finite precision $p$, absorbing constant base-conversion factors into $p$; for node-level prediction, $\vec{\theta}_{\mathrm{head}}$ denotes the parameters of the node output head, and $|\vec{\theta}_{\mathrm{head}}|$ denotes their number.

\paragraph{Node-level prediction.}
For node-level binary labels, write $\{(G^{(i)},\vec{y}^{(i)})\}_{i=1}^N\sim\D^N$, where $V_{\mathrm{variable}}(G)\subseteq V(G)$ denotes the ILP variable nodes to be predicted and $\vec{y}^{(i)}=(y_v^{(i)})_{v\in V_{\mathrm{variable}}(G^{(i)})}\in\{0,1\}^{V_{\mathrm{variable}}(G^{(i)})}$. 
Consider
{\small
\[
\begin{aligned}
\H^\mathcal{C}_{\mathrm{node}}
:=\{h:(G,v)\rightarrow\{0,1\}\mid
h \text{ is a }\colorgnn \text{ colored by }\mathcal{C}
\text{ and with a shared node-level output head}\}.
\end{aligned}
\]
}
For $h\in\H^C_{\mathrm{node}}$, define the node-level population and empirical risks
\[
L_{\mathrm{node}}(h):=\mathbb{E}_{(G,\vec{y})\sim\D}\left[\frac{1}{|V_{\mathrm{variable}}(G)|}\sum_{v\in V_{\mathrm{variable}}(G)}\mathbf{1}\{h(G,v)\neq y_v\}\right],
\]
and
\[
\hat{L}_{\mathrm{node}}(h):=\frac{1}{N}\sum_{i=1}^N\frac{1}{|V_{\mathrm{variable}}(G^{(i)})|}\sum_{v\in V_{\mathrm{variable}}(G^{(i)})}\mathbf{1}\{h(G^{(i)},v)\neq y_v^{(i)}\}.
\]
We write $\gen_{\mathrm{node}}(h,\D,N):=L_{\mathrm{node}}(h)-\hat{L}_{\mathrm{node}}(h)$.

\begin{theorem}[Node-level generalization]\label{thm:node-gen}
Fix a $d$-hop unique coloring rule $\mathcal{C}$ and consider the hypothesis class $\H^\mathcal{C}_{\mathrm{node}}$ realized by a depth-$d$ $\colorgnn$ with a shared node-level output head. Then, for any $\delta\in(0,1)$, with probability at least $1-\delta$ over $Z\sim\D^N$, every $h\in\H^\mathcal{C}_{\mathrm{node}}$ satisfies
\[
\big|\gen_{\mathrm{node}}(h,\D,N)\big|
\leq
\sqrt{\frac{p\cdot (|\mathcal{C}|\cdot |\vec{\theta}_{\emb}|+d\cdot |\vec{\theta}_{\merge}|+|\vec{\theta}_{\mathrm{head}}|)+\log(2/\delta)}{2N}}.
\]
\end{theorem}
The proof is provided in Appendix~\ref{app:node-gene}.  Consequently, to guarantee a node-level generalization gap of at most $\epsilon$ with probability at least $1-\delta$, it suffices to choose
\[
N\geq\frac{p\cdot (|\mathcal{C}|\cdot |\vec{\theta}_{\emb}|+d\cdot |\vec{\theta}_{\merge}|+|\vec{\theta}_{\mathrm{head}}|)+\log(2/\delta)}{2\epsilon^2}.
\]

\paragraph{Graph-level prediction.}
For graph-level binary labels, write $Z=\{(G^{(i)},y^{(i)})\}_{i=1}^N\sim\D^N$ with $y^{(i)}\in\{0,1\}$. Consider
\[
\H^\mathcal{C}_{\mathrm{graph}}:=\{h:G\rightarrow\{0,1\}\mid h \text{ is a }\colorgnn \text{ colored by }\mathcal{C}\text{ with fixed parameter-free graph readout}\}.
\]
For $h\in\H^\mathcal{C}_{\mathrm{graph}}$, define the graph-level population and empirical risks
\[
L_{\mathrm{graph}}(h):=\Pr_{(G,y)\sim \D}[h(G)\neq y],\qquad
\hat{L}_{\mathrm{graph}}(h):=\frac{1}{N}\sum_{i=1}^N \mathbf{1}\{h(G^{(i)})\neq y^{(i)}\},
\]
and the generalization gap $\gen_{\mathrm{graph}}(h,\D,N):=L_{\mathrm{graph}}(h)-\hat{L}_{\mathrm{graph}}(h)$.

\begin{theorem}[Graph-level generalization]\label{thm:gen}
Fix a $d$-hop unique coloring rule $\mathcal{C}$ and consider the hypothesis class $\H^\mathcal{C}_{\mathrm{graph}}$ realized by a depth-$d$ $\colorgnn$ with fixed parameter-free graph readout. Then, for any $\delta\in(0,1)$, with probability at least $1-\delta$ over $Z\sim\D^N$, every $h\in\H^\mathcal{C}_{\mathrm{graph}}$ satisfies
\[
\big|\gen_{\mathrm{graph}}(h,\D,N)\big|\leq \sqrt{\frac{p\cdot (|\mathcal{C}|\cdot |\vec{\theta}_{\emb}|+d\cdot |\vec{\theta}_{\merge}|)+\log(2/\delta)}{2N}}.
\]
\end{theorem}
The proof is provided in Appendix~\ref{app:gene}. Consequently, to guarantee a graph-level generalization gap of at most $\epsilon$ with probability at least $1-\delta$, it suffices to choose
\[
N\geq\frac{p\cdot(|\mathcal{C}|\cdot |\vec{\theta}_{\emb}|+d\cdot |\vec{\theta}_{\merge}|)+\log(2/\delta)}{2\epsilon^2}.
\]

\section{Experiments}
We organize the experiments into three parts: (i) setup, specifying the evaluation protocol and baseline methods; (ii) main experiments, validating the proposed scheme; and (iii) ablation studies, providing empirical analysis of the method and further extensions. The source code is available at \url{https://github.com/machinaddiffis/dhop_Coloring4GNN}.

{{The appendix provides additional experimental details and analyses: Appendix~\ref{app:dhop-performance} reports the BPP $d$-hop coloring ablation;} Appendix~\ref{app:experimental-setup-details} gives more detailed model architectures and training protocol; Appendix~\ref{app:identifier-shortcut} verifies the ID-shortcut phenomenon; Appendix~\ref{app:weak-coloring} compares our Local-UID coloring with another coloring method; Appendix~\ref{app:stronger-backbone-check} ablates the effect of stronger backbones; Appendix~\ref{app:train-test-losses} reports training and test losses across the full experimental suite as a proxy for fitting and expressive power; and Appendix~\ref{app:coloring-time} reports the coloring-time overhead.}

\paragraph{Benchmarks} We evaluate three ILP benchmarks that exhibit significant symmetry and two general ILP benchmarks: (i) Bin Packing Problem (BPP; \citep{schwerin1997bin}), 500 instances, 420 variables and 40 constraints per instance; (ii) Balanced Item Placement (BIP; \citep{gasse2022machine}), 300 instances, 1083 variables and 195 constraints;  (iii) Steel Mill Slab Design Problem (SMSP; \citep{schaus2011solving}), 380 instances, 22000--24000 variables and $\sim$10000 constraints; (iv) Independent Set (IS; \citep{gasse2019exact}), 400 instances, 1500 variables and 600 constraints; (v) Combinatorial Auction (CA; \citep{gasse2019exact}), 400 instances, 1500 variables and 6396 constraints.

\paragraph{Baselines}
(i) \textbf{No-Aug}: A baseline with no feature augmentation.
(ii) \textbf{Position}~(Global-UID): A feature-augmentation scheme that assigns distinct numbers to variable nodes~\citep{han2023gnn, chen2024symilo}.
(iii) \textbf{Uniform}~(Global-UID): A feature-augmentation scheme that adds i.i.d.\ noise sampled from a uniform distribution to every node~\citep{Chen0WY23a}.
(iv) \textbf{Orbit}: A feature-augmentation scheme that assigns UIDs based on symmetry orbits~\citep{chen2025gnns}.
(v) \textbf{Orbit+}: An extension of Orbit that additionally leverages inter-orbit relations~\citep{chen2025gnns}.

\paragraph{Metrics} For ILPs, following \citep{chen2025gnns}, we adopt \emph{Top-m\% error} as the metric. {For all five benchmarks (BPP, BIP, SMSP, IS, and CA), predictions are evaluated against binary decision vectors, and $\mathrm{Round}$ denotes the same coordinate-wise thresholding for every method: $\mathrm{Round}(\hat{y}_i)=\mathbf{1}\{\hat{y}_i>0.5\}$.} Specifically, given the ground-truth $y$ and a prediction $\hat{y}$, define the closest symmetric equivalent of $y$ as $\tilde{y}=\pi^*(y)$ with $\pi^*=\arg\min_{\pi}\|\hat{y}-\pi(y)\|_1$. {Let $S(m)$ be the index set of the top $m\%$ variables with the smallest values of $|\mathrm{Round}(\hat{y}_j)-\tilde{y}_j|$. The Top-$m\%$ error is defined as $\mathcal{E}(m)=\sum_{i\in S(m)}\left|\mathrm{Round}(\hat{y}_i)-\tilde{y}_i\right|$.}

\subsection{Main Experiments}
Table~\ref{tab:maj1} reports the main comparison across five ILP benchmarks. We separate the discussion into two regimes. The first three datasets (BPP, BIP, SMSP) are symmetry-heavy ILP benchmarks considered by \citet{chen2025gnns}. The last two datasets (IS, CA) test whether the same $\colorgnn$ and lightweight $\colorlui$ designs transfer to more general ILP benchmarks.

\paragraph{Symmetry-heavy ILP benchmarks}
For BPP/BIP/SMSP, Table~\ref{tab:maj1} empirically validates the effectiveness of the $\colorgnn$ architecture in four respects. (i) \textbf{Effectiveness of color-aware embedding.} $\colorgnn$ consistently improves over the Global-UID baselines (Non-Aug, Uniform, Position) and outperforms Orbit/Orbit+ on most reported metrics. (ii) \textbf{Lightweight variant.} The special case $\colorlui$ achieves performance comparable to $\colorgnn$, making it a practical lightweight substitute for large-scale graphs. (iii) \textbf{Exception on SMSP.} On Top-30\% and Top-50\% for SMSP, Orbit and Orbit+ perform better. This likely stems from the fact that Orbit incorporates instance-specific information by computing solution orbits and leverages an evaluation metric that normalizes predictions under these symmetries, whereas our approach derives enhanced representational power solely from the underlying graph structure, without relying on such instance-level preprocessing. (iv) \textbf{Compatibility with prior UID schemes.} Augmenting $\colorgnn$ with Orbit or Orbit+ yields substantial additional gains: $\colorgnn(\aggr)$ with Orbit+ improves the BPP Top-70\% and Top-100\% errors to $0.02$ and $12.16$, respectively.

\paragraph{Transfer to broader ILP benchmarks}
For IS/CA, Table~\ref{tab:maj1} shows that the Local-UID scheme transfers directly to broader ILP benchmarks and achieves the best reported performance under the generic-baseline protocol. The improvement is especially strong on IS, where both $\colorgnn$ and $\colorlui$ perform well across the reported metrics. On CA, $\colorgnn$ is best throughout, and $\colorlui$ also consistently improves over Non-Aug, Uniform, and Position. {Orbit defines UID classes from formulation-symmetry orbits, and Orbit+ additionally uses valid inter-orbit correspondences~\citep{chen2025gnns}, which are problem-structure dependent. We therefore mark Orbit/Orbit+ as N/A in Table~\ref{tab:ab2} and omit them from the IS/CA blocks in Table~\ref{tab:maj1}.} In contrast, Local-UIDs require only the graph coloring construction and can be applied directly to IS and CA without problem-specific orbit preprocessing.

\begin{table*}[t]
\centering
\caption{Comparative results on all datasets for our methods and baselines, evaluated across Top-$m\%$ thresholds (lower is better).}
\small
\resizebox{0.98\linewidth}{!}{
\begin{tabular}{llccccc}
\toprule
\textbf{Dataset} & \textbf{Methods} & \textbf{30\%} & \textbf{50\%} & \textbf{70\%} & \textbf{90\%} & \textbf{100\%} \\
\midrule
\multirow{7}{*}{BPP}
& Non-Aug & $6.31{\pm}0.04$ & $10.47{\pm}0.03$ & $14.62{\pm}0.03$ & $18.76{\pm}0.06$ & $28.87{\pm}0.00$ \\
& Position & $0.34{\pm}0.36$ & $1.26{\pm}1.46$ & $3.05{\pm}2.65$ & $8.70{\pm}2.42$ & $19.69{\pm}1.86$ \\
& Uniform & $0.56{\pm}0.43$ & $2.27{\pm}1.73$ & $4.65{\pm}2.99$ & $10.48{\pm}1.57$ & $20.78{\pm}0.31$ \\
& Orbit & \best{$0.00{\pm}0.00$} & $0.01{\pm}0.01$ & $0.03{\pm}0.02$ & $3.36{\pm}0.14$ & $14.55{\pm}0.39$ \\
& Orbit+ & \best{$0.00{\pm}0.00$} & $0.01{\pm}0.00$ & \best{$0.03{\pm}0.01$} & $3.02{\pm}0.08$ & $13.51{\pm}0.22$ \\
\cmidrule(lr){2-7}
& $\colorlui$ & \best{$0.00{\pm}0.00$} & $0.03{\pm}0.02$ & $0.18{\pm}0.07$ & $2.66{\pm}0.45$ & \best{$12.52{\pm}0.98$} \\
& $\colorgnn$ & \best{$0.00{\pm}0.00$} & \best{$0.00{\pm}0.00$} & $0.09{\pm}0.04$ & \best{$2.40{\pm}0.40$} & $13.13{\pm}1.01$ \\
\midrule
\multirow{7}{*}{BIP}
& Non-Aug & $29.23{\pm}3.74$ & $50.18{\pm}4.15$ & $71.51{\pm}3.46$ & $93.89{\pm}0.72$ & $105.00{\pm}0.00$ \\
& Uniform & $4.55{\pm}0.42$ & $14.30{\pm}1.36$ & $44.78{\pm}1.31$ & $80.54{\pm}1.63$ & $104.86{\pm}0.29$ \\
& Position & $3.99{\pm}0.23$ & $12.24{\pm}1.22$ & $45.02{\pm}1.25$ & $81.78{\pm}0.53$ & $104.59{\pm}0.22$ \\
& Orbit & $3.20{\pm}0.12$ & $6.20{\pm}0.18$ & $37.02{\pm}0.87$ & $74.51{\pm}0.79$ & $102.23{\pm}0.15$ \\
& Orbit+ & $3.00{\pm}0.09$ & $5.33{\pm}0.13$ & $39.25{\pm}0.05$ & $79.30{\pm}0.07$ & $101.90{\pm}0.11$ \\
\cmidrule(lr){2-7}
& $\colorlui$ & $2.73{\pm}0.10$ & $5.09{\pm}0.13$ & $33.46{\pm}0.79$ & $73.08{\pm}0.84$ & $98.84{\pm}0.87$ \\
& $\colorgnn$ & \best{$0.244{\pm}0.072$} & \best{$3.850{\pm}0.247$} & \best{$32.221{\pm}1.690$} & \best{$72.360{\pm}1.524$} & \best{$97.666{\pm}0.274$} \\
\midrule
\multirow{7}{*}{SMSP}
& Non-Aug & $33.13{\pm}0.39$ & $56.87{\pm}0.77$ & $81.85{\pm}0.78$ & $109.34{\pm}0.24$ & $214.80{\pm}0.73$ \\
& Position & $0.12{\pm}0.04$ & $3.37{\pm}0.60$ & $26.99{\pm}0.71$ & $62.47{\pm}2.25$ & $146.69{\pm}1.04$ \\
& Uniform & $9.04{\pm}10.21$ & $18.33{\pm}17.16$ & $38.54{\pm}13.82$ & $68.36{\pm}7.91$ & $150.41{\pm}4.15$ \\
& Orbit & \best{$0.01{\pm}0.01$} & $0.83{\pm}0.65$ & $16.21{\pm}1.95$ & $50.77{\pm}1.00$ & $145.36{\pm}0.78$ \\
& Orbit+ & \best{$0.01{\pm}0.01$} & \best{$0.36{\pm}0.17$} & $11.91{\pm}0.94$ & $50.88{\pm}0.41$ & $138.37{\pm}1.01$ \\
\cmidrule(lr){2-7}
& $\colorlui$ & $0.23{\pm}0.08$ & $1.88{\pm}0.41$ & \best{$7.61{\pm}1.07$} & $17.72{\pm}1.33$ & $119.42{\pm}2.79$ \\
& $\colorgnn$ & $2.45{\pm}0.21$ & $4.94{\pm}0.38$ & $8.96{\pm}0.77$ & \best{$16.45{\pm}1.27$} & \best{$117.17{\pm}3.27$} \\
\midrule
\multirow{5}{*}{IS}
& Non-Aug & \best{$0.00{\pm}0.00$} & $43.49{\pm}12.50$ & $193.19{\pm}36.45$ & $299.50{\pm}13.29$ & $353.76{\pm}0.29$ \\
& Position & \best{$0.00{\pm}0.00$} & $0.79{\pm}0.16$ & $29.97{\pm}0.13$ & $132.93{\pm}0.64$ & $206.74{\pm}0.97$ \\
& Uniform & \best{$0.00{\pm}0.00$} & $0.89{\pm}0.13$ & $30.84{\pm}0.51$ & $134.68{\pm}1.20$ & $208.54{\pm}1.80$ \\
\cmidrule(lr){2-7}
& $\colorgnn$ & \best{$0.00{\pm}0.00$} & \best{$0.00{\pm}0.00$} & \best{$0.14{\pm}0.02$} & \best{$21.48{\pm}0.36$} & $74.01{\pm}0.51$ \\
& $\colorlui$ & \best{$0.00{\pm}0.00$} & \best{$0.00{\pm}0.00$} & $0.15{\pm}0.04$ & $21.68{\pm}0.41$ & \best{$73.93{\pm}0.61$} \\
\midrule
\multirow{5}{*}{CA}
& Non-Aug & $83.80{\pm}4.61$ & $140.90{\pm}6.41$ & $200.31{\pm}5.36$ & $261.27{\pm}1.13$ & $291.24{\pm}0.22$ \\
& Position & $34.32{\pm}0.43$ & $76.84{\pm}0.34$ & $136.86{\pm}0.42$ & $223.17{\pm}0.58$ & $286.42{\pm}0.71$ \\
& Uniform & $36.01{\pm}0.40$ & $79.30{\pm}0.40$ & $139.07{\pm}0.48$ & $224.84{\pm}0.65$ & $287.63{\pm}1.06$ \\
\cmidrule(lr){2-7}
& $\colorgnn$ & \best{$27.70{\pm}0.43$} & \best{$66.65{\pm}0.51$} & \best{$125.62{\pm}0.37$} & \best{$215.18{\pm}0.69$} & \best{$280.27{\pm}0.57$} \\
& $\colorlui$ & $29.86{\pm}0.54$ & $70.83{\pm}0.84$ & $131.13{\pm}0.66$ & $220.48{\pm}0.83$ & $284.95{\pm}0.79$ \\
\bottomrule
\end{tabular}
}
\label{tab:maj-main-all}
\label{tab:maj1}
\label{tab:maj2-bip}
\label{tab:maj2-smsp}
\label{tab:maj2-is-ca}
\end{table*}

\subsection{Ablation Studies}\label{sec:methodab}
{This section has two aims: (i) provide evidence that the observed gains are consistent with our feature-augmentation principle; and (ii) probe architectural sensitivity by varying the color-embedding function, suggesting that benefits arise from color integration rather than a specific functional choice. We report the parameter-$d$ ablation in Appendix~\ref{app:dhop-performance} (Table~\ref{tab:ab3}), which directly supports our theory.}

\paragraph{Sources of Performance Gains} (i) The gains of our methods are not due to increased dimensionality: On BPP, we compare $\colorgnn$ against Non-Aug and Non-Aug+ (which scales the baseline embedding dimension by a factor of $|C|$). We find that merely increasing model capacity without color information does not improve performance and can even degrade it (e.g., Non-Aug vs. Non-Aug+, see the upper half of Table \ref{tab:ab1}). (ii) $\colorgnn$ typically uses a smaller but sufficient number of colors: Table \ref{tab:ab2} shows that, our methods uses very few UIDs on BPP, SMSP,IS and CA. On BIP, our method uses more colors than the Orbit/Orbit+ schemes yet achieves better performance.
In BIP, a 1-hop unique coloring requires at least 106 colors, as there exists a constraint involving 106 variables. Consequently, Orbit and Orbit+, which use only 51 colors, lack sufficient expressive power to fully distinguish all 1-hop neighbors.
\begin{table*}[t]
\centering
\caption{{Comparative results on the BPP dataset for color-embedding variants, the Non-Aug baseline, and their combinations, evaluated across Top-m\% thresholds (lower is better).}}
\small
\setlength{\tabcolsep}{5pt}
\begin{tabular}{lrrrrr}
\toprule
\textbf{Method} & \textbf{Top-30\%} & \textbf{Top-50\%} & \textbf{Top-70\%} & \textbf{Top-90\%} & \textbf{Top-100\%} \\
\midrule

Non-Aug                      & $6.31{\pm}0.04$ & $10.47{\pm}0.03$ & $14.62{\pm}0.03$ & $18.76{\pm}0.06$ & $28.87{\pm}0.00$ \\
Non-Aug+                     & $6.35{\pm}0.03$ & $10.44{\pm}0.01$ & $14.62{\pm}0.02$ & $18.82{\pm}0.04$ & $28.87{\pm}0.00$ \\
Orbit                        & \best{$0.00{\pm}0.00$} & $0.01{\pm}0.01$ & $0.03{\pm}0.02$ & $3.36{\pm}0.14$ & $14.55{\pm}0.39$ \\
Orbit+                       & \best{$0.00{\pm}0.00$} & \best{$0.01{\pm}0.00$} & \best{$0.03{\pm}0.01$} & \best{$3.02{\pm}0.08$} & \best{$13.51{\pm}0.22$} \\
\midrule
$\colorgnn$                  & \best{$0.00{\pm}0.00$} & \best{$0.00{\pm}0.00$} & \best{$0.09{\pm}0.04$} & \best{$2.40{\pm}0.40$} & \best{$13.13{\pm}1.01$} \\
$\colorgnn(\merge)$          & $0.01{\pm}0.00$ & $0.02{\pm}0.00$ & $0.33{\pm}0.11$ & $4.93{\pm}1.82$ & $16.53{\pm}2.49$ \\
$\colorgnn(\aggr)$           & \best{$0.00{\pm}0.00$} & \best{$0.00{\pm}0.00$} & $0.29{\pm}0.06$ & $4.69{\pm}0.51$ & $16.70{\pm}0.81$ \\
\midrule
$\colorgnn$,Orbit            & $0.01{\pm}0.01$ & $0.07{\pm}0.04$ & $0.59{\pm}0.15$ & $5.60{\pm}1.75$ & $16.30{\pm}2.89$ \\
$\colorgnn(\merge)$,Orbit    & \best{$0.00{\pm}0.00$} & $0.03{\pm}0.03$ & $0.61{\pm}0.15$ & $6.61{\pm}0.35$ & $17.89{\pm}0.84$ \\
$\colorgnn(\aggr)$,Orbit     & \best{$0.00{\pm}0.00$} & \best{$0.01{\pm}0.01$} & \best{$0.02{\pm}0.01$} & \best{$2.96{\pm}0.39$} & \best{$13.22{\pm}1.18$} \\
\midrule
$\colorgnn$,Orbit+           & \best{$0.00{\pm}0.00$} & $0.01{\pm}0.01$ & $0.35{\pm}0.16$ & $3.55{\pm}0.59$ & $13.31{\pm}0.93$ \\
$\colorgnn(\merge)$,Orbit+   & \best{$0.00{\pm}0.00$} & $0.03{\pm}0.01$ & $0.47{\pm}0.09$ & $4.28{\pm}0.53$ & $14.59{\pm}0.78$ \\
$\colorgnn(\aggr)$,Orbit+    & \best{$0.00{\pm}0.00$} & \best{$0.00{\pm}0.00$} & \best{$0.02{\pm}0.01$} & \best{$2.59{\pm}0.08$} & \best{$12.16{\pm}0.25$} \\
\bottomrule
\end{tabular}
\label{tab:ab1}
\end{table*}
\vspace{-1ex}

\begin{table}[t]
\centering
\caption{Number of unique identifiers (UIDs) used by each method across five benchmark datasets.}
\small
\setlength{\tabcolsep}{7pt}
\begin{tabular}{lrrrrr}
\toprule
\textbf{Method} & \textbf{BPP} & \textbf{BIP} & \textbf{SMSP} & \textbf{IS} & \textbf{CA} \\
\midrule
Uniform & 420 & 1082.41 & 23137.25 & 1498.89 & 1498.89 \\
Position       & 420 & 1083 & 23417.97 & 1500.00 & 1500.00 \\
Orbit          & 140 & \textbf{51} & 1000 & N/A & N/A \\
Orbit+         & 140 & \textbf{51} & 6070 & N/A & N/A \\
\midrule
$\colorgnn$       & \textbf{32} & 114 & \textbf{171.68} & \textbf{4.27} & \textbf{13.25} \\
\bottomrule
\end{tabular}
\label{tab:ab2}
\end{table}

{
\paragraph{Coloring Schemes with Alternative Functional Forms}
Embedding structural UID information into the GNN architecture improves performance (Table~\ref{tab:ab1}).
We study three $\colorgnn$ variants, which are parallel architectural choices that differ only in \emph{where} the color signal is injected into message passing. Let $C(v)$ denote the color of node $v$. In the message-passing step, incorporating color into the aggregation operator yields $\colorgnn(\aggr)$:
{\[
h_{v}^{(k)} :=
\merge^{(k)}\!\left(
h_{v}^{(k-1)},
\aggr\bigl(\{(\alpha_{C(u)} \cdot h_{u}^{(k-1)},\, W(u,v))\mid u\in N_1(v)\})
\right),
\]}
where $\alpha_{C(u)}\in\mathbb{R}$ is a learnable scalar associated with color $C(u)$, initialized as $\alpha_{C(u)} \sim \mathcal N(1,1)$ for all $u$. Alternatively, injecting color into the merge function yields $\colorgnn(\merge)$:

\begin{align*}
h_v^{(k)} &:= \merge^{(k)}_{C(v)}\!\Bigl(h_v^{(k-1)},\, \aggr({\{(\vec{h}^{(k-1)}_{u},W(u,v))\mid u\in N_1(v)\}})\Bigr).
\end{align*}
where $\merge^{(k)}_{C(v)}$ is a learnable mapping specific to color $C(v)$.The embedding-level $\colorgnn$ is the strongest standalone variant on BPP. When combined with Orbit+, $\colorgnn(\aggr)$ achieves the best performance in Table~\ref{tab:ab1}.This points to a promising direction for future work: matching different color-injection variants with complementary augmentation or UID schemes on other tasks may yield further gains.
}

\FloatBarrier
\section{Conclusion and Limitations}

This paper introduced a parsimonious feature-augmentation scheme for GNNs on ILPs based on $d$-hop unique coloring, along with a novel color-aware architecture $\colorgnn$ and its lightweight variant $\colorlui$. Theoretically, we proved that compared with conventional Global-UID schemes, our Local-UID approach achieves the same expressive power while offering superior generalization. 
{Extensive experiments validate its effectiveness across diverse benchmarks, including both symmetry-heavy ILP tasks and broader ILP benchmarks. The ablation studies further show that Local-UIDs can be effectively combined with other UID schemes. Additional appendix results provide theory-consistent evidence on the role of the coloring radius and show that the benefits persist under stronger backbones.}

Our results have several limitations. First, the theoretical expressivity guarantee requires the coloring radius to match the GNN receptive field to recover the expressive power of Global-UIDs. In practice, however, one may choose a smaller coloring radius to reduce preprocessing cost and the number of colors; this empirical tradeoff is not fully characterized by our theory. Second, although Local-UIDs substantially reduce the number of identifiers compared with Global-UIDs, the number of colors can still grow with the $2d$-hop degree. Thus, for dense graphs or large coloring radius, the preprocessing cost and the parameter cost of $\colorgnn$ may become significant.

\bibliographystyle{plainnat}
\bibliography{example_paper}

\appendix

\section{Related Works}
{The following positional-encoding (PE) methods are \emph{complementary} to our color-based UID scheme and are therefore best viewed as orthogonal design dimensions rather than head-to-head baselines. Graph Transformers with Laplacian-eigenvector absolute positional encodings (LapPE) were introduced by \citet{dwivedi2020generalization}. \citet{lim2022sign} design architectures that are invariant to eigenvector sign and eigenspace basis, making spectral PEs robust to symmetry choices. \citet{huang2023stability} further improve stability and expressivity by processing eigenvectors via eigenvalue-weighted, softly partitioned eigenspaces. Beyond explicit eigenvectors, \citet{eliasof2023graph} obtain scalable PEs by filtering random signals with the graph Laplacian, while \citet{kanatsoulis2025learning} propose permutation-invariant features with lower computational cost and improved stability. \citet{dwivedi2021graph} decouple random-walk–based PEs from other structural signals to build more flexible architectures, and \citet{bechler2024utilization} align latent spaces induced by different PEs via tailored losses to promote permutation equivariance. 

Structural Message Passing (SMP)~\citep{vignac2020building} enhances expressive power by propagating a local-context matrix initialized with one-hot node identifiers. This perspective is complementary to ours: one can, for example, replace the SMP context matrix with our color features, or incorporate their node identifiers into our $\colorgnn$ architecture. Under the distributed local algorithm model, \cite{sato2019approximation} prove that weak 2-coloring strictly increases the capability of GNNs. In a related vein, \cite{sato2021random} use random node features that act as global UIDs to enhance GNNs. Our work generalizes this line of research by showing that a parsimonious family of \emph{local} UID schemes (such as $d$-hop colorings) suffices to significantly improve GNN expressivity across a wide range of combinatorial and general-graph tasks. Complementary to these perspectives, \cite{grotschla2024benchmarking} provide a systematic benchmark of PEs for both message-passing GNNs and graph Transformers\citep{dwivedi2020generalization}. In their framework, PEs are treated as node-feature augmentations that encode graph topology and can be plugged into arbitrary GNN/GT backbones. Our $\colorgnn$ fits naturally into this view: it can be interpreted as a new family of structural positional encodings based on locally unique colorings (e.g., $d$-hop unique coloring) that provide symmetry-breaking identity information while preserving permutation equivariance. }

\section{Proofs}
\label{all_proof}
{
\subsection{Proof of Proposition \ref{prop:dhop1}}
\begin{proof}
(\emph{$\Rightarrow$}) Suppose the coloring is $d$-hop unique. Take any two vertices $u,w$ with $\mathrm{dist}(u,w)\leq 2d$. A shortest $u$-$w$ path has a midpoint of $v$ with both 
$\mathrm{dist}(u,v),\mathrm{dist}(w,v)\le d$,
so $u, w$ are in $N_d(v)\cup\{v\}$ and must have distinct colors.
(\emph{$\Leftarrow$}) Suppose the coloring is a $(2d)$-hop coloring. Then for any $v$ and any distinct $u,w\in N_d(v)\cup\{v\}$, we have 
$\mathrm{dist}(u,w)\le \mathrm{dist}(u,v)+\mathrm{dist}(v,w)\le d+d=2d$
by the triangle inequality. So $u$ and $w$ must have different colors. Thus each vertex in $N_d(v)\cup\{v\}$ is uniquely colored.
\end{proof}
\subsection{Proof of Lemma \ref{prop:gnn=lui}}
\begin{proof}
Fix a coloring $C$. On the one hand, both $\colorgnn$ and $\colorlui$ with $C$ can be simulated by a $d$-round distributed algorithm with the same coloring: one layer corresponds to one round, and the coloring $C$ is provided as auxiliary input.

On the other hand, given any $d$-round distributed algorithm with coloring $C$, we can construct $d$-layer $\colorgnn$ and $\colorlui$ that faithfully simulate it. For $\colorlui$, by the universal approximation property of MLP, the merge function $\merge^{(k)}$ can be parameterized to implement the local computation in round $k$. For $\colorgnn$, we use the embedding mapping {$\emb_c(\vec{v})=(c,\vec{v})$} to preserve the coloring, and similarly choose the merge function $\merge^{(k)}$ to implement the local computations, with $C$ explicitly incorporated as input features. Now, the lemma follows.
\end{proof}
\subsection{Proof of Theorem \ref{thm:d-hop}}
\begin{proof}
By Lemma \ref{prop:gnn=lui}, it suffices to show that a $d$-round distributed algorithm with a $d$-hop unique coloring is as powerful as a $d$-round distributed algorithm with global unique identifiers. In a $d$-round distributed algorithm with global unique identifiers, each node $v$ bases its output solely on the induced $d$-hop subgraph {$B_d(v)$}. A $d$-hop unique coloring ensures that all nodes in {$B_d(v)$} receive distinct colors; thus the induced subgraph {$B_d(v)$} can be reconstructed by the node $v$. 
Now, the theorem is immediate.
\end{proof}

}

\subsection{\texorpdfstring{Proof of Theorem~\ref{thm:node-gen}}{Proof of Theorem}}\label{app:node-gene}
\begin{proof}
We first prove a generic finite-class bounded-loss inequality. This argument is independent of the node-level setting and will also be used in the proof of Theorem~\ref{thm:gen}. Let $H$ be any finite hypothesis class and let $S_1,\ldots,S_N$ be i.i.d. samples. Suppose each $h\in H$ has a loss $\ell_h(S)\in[0,1]$, and define
\[
L_{\ell}(h):=\mathbb{E}[\ell_h(S)],
\qquad
\hat{L}_{\ell}(h):=\frac{1}{N}\sum_{i=1}^N \ell_h(S_i).
\]
For a fixed $h$, Hoeffding's inequality gives, for every $\epsilon>0$,
\[
\Pr\left(\big|L_{\ell}(h)-\hat{L}_{\ell}(h)\big|>\epsilon\right)
\leq 2\exp(-2N\epsilon^2).
\]
Taking a union bound over $H$ yields
\[
\Pr\left(\exists h\in H:\big|L_{\ell}(h)-\hat{L}_{\ell}(h)\big|>\epsilon\right)
\leq 2|H|\exp(-2N\epsilon^2).
\]
Choosing
\[
\epsilon=\sqrt{\frac{\log(2|H|)+\log(1/\delta)}{2N}}
\]
makes the right-hand side at most $\delta$. Hence, with probability at least $1-\delta$,
\begin{equation}
\label{eq:finite-class-bound}
\sup_{h\in H}\big|L_{\ell}(h)-\hat{L}_{\ell}(h)\big|
\leq
\sqrt{\frac{\log(2|H|)+\log(1/\delta)}{2N}}.
\end{equation}

We now reduce the node-level risk to this bounded-loss setting. Let $Z=\{(G^{(i)},\vec{y}^{(i)})\}_{i=1}^N\sim\D^N$, where $\vec{y}^{(i)}=(y_v^{(i)})_{v\in V_{\mathrm{variable}}(G^{(i)})}\in\{0,1\}^{V_{\mathrm{variable}}(G^{(i)})}$. Fix $h\in\H^C_{\mathrm{node}}$. For the $i$-th labeled graph sample, define the averaged node error
\[
\xi_i(h):=\frac{1}{|V_{\mathrm{variable}}(G^{(i)})|}\sum_{v\in V_{\mathrm{variable}}(G^{(i)})}\mathbf{1}\{h(G^{(i)},v)\neq y_v^{(i)}\}.
\]
The denominator is well defined for the ILP node-prediction task, where every sampled graph has at least one variable node to be predicted. Each summand is an indicator, so $0\le \xi_i(h)\le 1$. Since the labeled graph samples $(G^{(1)},\vec{y}^{(1)}),\ldots,(G^{(N)},\vec{y}^{(N)})$ are i.i.d. samples from $\D$, the random variables $\xi_1(h),\ldots,\xi_N(h)$ are also i.i.d. No independence among node labels inside the same graph is required, because $\xi_i(h)$ is treated as one bounded loss for the whole labeled graph sample. Moreover, by the definitions of the node-level population and empirical risks in Section~\ref{sec:generalization},
\[
\mathbb{E}[\xi_i(h)]=L_{\mathrm{node}}(h),\qquad
\frac{1}{N}\sum_{i=1}^N\xi_i(h)=\hat{L}_{\mathrm{node}}(h).
\]
Applying~\eqref{eq:finite-class-bound} with $H=\H^\mathcal{C}_{\mathrm{node}}$ and the sample-level loss
\[
\ell_h((G,\vec{y})):=\frac{1}{|V_{\mathrm{variable}}(G)|}\sum_{v\in V_{\mathrm{variable}}(G)}\mathbf{1}\{h(G,v)\neq y_v\}
\]
gives, with probability at least $1-\delta$,
\[
\sup_{h\in\H^\mathcal{C}_{\mathrm{node}}}\big|L_{\mathrm{node}}(h)-\hat{L}_{\mathrm{node}}(h)\big|
\leq
\sqrt{\frac{\log(2|\H^C_{\mathrm{node}}|)+\log(1/\delta)}{2N}}.
\]

It remains to bound the size of $\H^\mathcal{C}_{\mathrm{node}}$. Compared with the graph-level class, the node-level class adds a node output head with parameters $\vec{\theta}_{\mathrm{head}}$, so the total number of trainable parameter entries is at most
\[
M_{\mathrm{node}}
:=
|\mathcal{C}|\cdot|\vec{\theta}_{\emb}|+d\cdot|\vec{\theta}_{\merge}|+|\vec{\theta}_{\mathrm{head}}|.
\]
The finite-precision convention in Section~\ref{sec:generalization} yields
\[
\log|\H^\mathcal{C}_{\mathrm{node}}|
\leq pM_{\mathrm{node}}
=p\cdot\bigl(|\mathcal{C}|\cdot|\vec{\theta}_{\emb}|+d\cdot|\vec{\theta}_{\merge}|+|\vec{\theta}_{\mathrm{head}}|\bigr).
\]
Substituting this into the preceding finite-class bound gives
\[
\sup_{h\in\H^\mathcal{C}_{\mathrm{node}}}\big|\gen_{\mathrm{node}}(h,\D,N)\big|
\leq
\sqrt{\frac{p\cdot\bigl(|\mathcal{C}|\cdot|\vec{\theta}_{\emb}|+d\cdot|\vec{\theta}_{\merge}|+|\vec{\theta}_{\mathrm{head}}|\bigr)+\log(2/\delta)}{2N}},
\]
which implies the claimed node-level uniform generalization bound.

\end{proof}

\subsection{\texorpdfstring{Proof of Theorem~\ref{thm:gen}}{Proof of Theorem}}\label{app:gene}
\begin{proof}
We first bound the size of the graph-level hypothesis class. The coloring rule $\mathcal{C}$ is fixed before sampling and label-independent; after its fixed deterministic color-name convention, all graphs use a common palette of size $|\mathcal{C}|$. The embedding stage therefore contains one color-specific embedding map per color, contributing
\[
|\mathcal{C}|\cdot|\vec{\theta}_{\emb}|
\]
trainable parameter entries. The depth-$d$ message-passing part contributes at most $|\vec{\theta}_{\merge}|$ parameter entries per layer, hence at most
\[
d\cdot|\vec{\theta}_{\merge}|
\]
parameter entries in total. The graph readout is fixed and parameter-free, so no readout parameters enter this count. Thus the total number of trainable parameter entries is at most
\[
M_{\mathrm{graph}}
:=
|\mathcal{C}|\cdot|\vec{\theta}_{\emb}|+d\cdot|\vec{\theta}_{\merge}|.
\]
By the finite-precision convention in Section~\ref{sec:generalization}, there are at most $\exp(pM_{\mathrm{graph}})$ parameter configurations; different configurations may induce the same binary classifier, so
\[
\log|\H^\mathcal{C}_{\mathrm{graph}}|
\leq pM_{\mathrm{graph}}
=p\cdot\bigl(|\mathcal{C}|\cdot|\vec{\theta}_{\emb}|+d\cdot|\vec{\theta}_{\merge}|\bigr).
\]

We now apply the generic finite-class inequality~\eqref{eq:finite-class-bound} established at the start of Appendix~\ref{app:node-gene}. Let $Z=\{(G^{(i)},y^{(i)})\}_{i=1}^N\sim\D^N$ be i.i.d. labeled samples. For a fixed $h\in\H^\mathcal{C}_{\mathrm{graph}}$, define
\[
X_i(h):=\mathbf{1}\{h(G^{(i)})\neq y^{(i)}\}.
\]
Then $0\le X_i(h)\le 1$, and the variables $X_1(h),\ldots,X_N(h)$ are i.i.d. By the definitions of the graph-level population and empirical risks in Section~\ref{sec:generalization},
\[
L_{\mathrm{graph}}(h)=\mathbb{E}[X_i(h)],
\qquad
\hat{L}_{\mathrm{graph}}(h)=\frac{1}{N}\sum_{i=1}^N X_i(h).
\]
Thus \eqref{eq:finite-class-bound}, applied with $H=\H^\mathcal{C}_{\mathrm{graph}}$ and the graph-level loss $\ell_h((G,y)):=\mathbf{1}\{h(G)\neq y\}$, gives with probability at least $1-\delta$ that
\[
\sup_{h\in\H^\mathcal{C}_{\mathrm{graph}}}\big|L_{\mathrm{graph}}(h)-\hat{L}_{\mathrm{graph}}(h)\big|
\leq
\sqrt{\frac{\log(2|\H^\mathcal{C}_{\mathrm{graph}}|)+\log(1/\delta)}{2N}}.
\]
Using the bound on $\log|\H^C_{\mathrm{graph}}|$ gives
\[
\sup_{h\in\H^\mathcal{C}_{\mathrm{graph}}}\big|\gen_{\mathrm{graph}}(h,\D,N)\big|
\leq
\sqrt{\frac{p\cdot\bigl(|\mathcal{C}|\cdot|\vec{\theta}_{\emb}|+d\cdot|\vec{\theta}_{\merge}|\bigr)+\log(2/\delta)}{2N}},
\]
which implies the claimed bound for every $h\in\H^\mathcal{C}_{\mathrm{graph}}$ and proves Theorem~\ref{thm:gen}.
\end{proof}

\section{Experimental Setup Details}
\label{app:experimental-setup-details}

\paragraph{Model architectures}
The goal is to learn a \(f\!: G \to \mathbb{R}^n\) to predict the optimal integer solution vector. Following \cite{chen2025gnns}, we use four “half-layer” graph-convolution blocks to extract hidden representations, followed by a two-layer MLP head for prediction.

\paragraph{Training protocol}
We use the unmodified training configurations from \citep{chen2025gnns,li2024pdhg} for the respective architectures, including learning rate, batch size, epochs, optimizer, weight decay, and scheduler~(see the original papers for specifications). {By Remark \ref{remark:radius},} 
we use greedy 1-hop unique coloring in the experiments.
We conduct all experiments under the same configuration on a server equipped with two Intel Xeon Gold 5117 CPUs (2.0 GHz), 256 GB RAM, and two NVIDIA V100 GPUs.

\paragraph{Compute resources}
Our method settings follow \citet{chen2025gnns}, and all models are trained until convergence. On the hardware described above, training all models to convergence requires approximately 40 hours in total.

\section{Effect of the coloring raduis $r$ on performance.}
\label{app:dhop-performance}

{
We evaluate $\colorlui$ with different coloring raduis $r$ on the BPP dataset (Table \ref{tab:ab3}). The results indicate that: 
\begin{itemize}
\item  4-hop coloring yields the best or tied-best performance across all reported thresholds, which is consistent with our theory that a 2-hop unique coloring, equivalently 4-hop coloring, suffices 2-layer GNNs; 
\item 2-hop coloring is not optimal in terms of accuracy, but it offers a pragmatic compromise between computational cost and performance; 
\item As $r$ increases beyond 4, the performance starts to deteriorate. Based on these observations, our practical guideline for choosing $r$ is to set it such that $r \leq {2}\cdot\text{GNN depth}$.
\end{itemize}

\begin{table}[H]
\centering
\caption{Comparative results on the BPP dataset for different coloring radius $r$ ($\colorlui$ with a $r$-hop coloring), evaluated across Top-$m\%$ thresholds (lower is better).}
\small
\setlength{\tabcolsep}{3pt}
\begin{tabular}{lrrrrr}
\toprule
\best{$\#r$-hop coloring} & \textbf{Top-30\%} & \textbf{Top-50\%} & \textbf{Top-70\%} & \textbf{Top-90\%} & \textbf{Top-100\%} \\
\midrule
1 & $6.39{\pm}0.02$ & $10.49{\pm}0.02$ & $14.62{\pm}0.02$ & $18.74{\pm}0.02$ & $28.87{\pm}0.00$ \\
2 (1-hop unique coloring)& \best{$0.00{\pm}0.00$} & $0.03{\pm}0.02$ & $0.18{\pm}0.07$ & $2.66{\pm}0.45$ & $12.52{\pm}0.98$ \\
3 & \best{$0.00{\pm}0.00$} & $0.02{\pm}0.01$ & $0.41{\pm}0.11$ & $5.84{\pm}1.82$ & $16.62{\pm}2.21$ \\
4 (2-hop unique coloring) & \best{$0.00{\pm}0.00$} & \best{$0.00{\pm}0.00$} & \best{$0.02{\pm}0.01$} & \best{$2.13{\pm}0.08$} & \best{$10.93{\pm}0.35$} \\
5 & \best{$0.00{\pm}0.00$} & \best{$0.00{\pm}0.00$} & $0.03{\pm}0.01$ & $2.23{\pm}0.08$ & $11.18{\pm}0.10$ \\
6 (3-hop unique coloring) & \best{$0.00{\pm}0.00$} & \best{$0.00{\pm}0.00$} & $0.03{\pm}0.01$ & $2.23{\pm}0.07$ & $11.32{\pm}0.22$ \\
\bottomrule
\end{tabular}
\label{tab:ab3}
\end{table}
}

\section{{Synthetic ID-Permutation Test for Identifier Shortcuts}}
\label{app:identifier-shortcut}

We conduct a synthetic experiment to isolate whether Global-UID-style identifiers can induce shortcut reliance. The experiment is a diagnostic rather than a benchmark for end-to-end ILP solving or feasibility prediction.

We construct ILP-style bipartite graphs for a query-variable binary classification task. Each graph contains a designated query variable $q$, and the model predicts the binary label of $q$. This setup gives a shallow GNN a local binary decision problem, mirroring the variable-prediction objective in our ILP tasks. The label is determined only by the local structure around the two constraint nodes adjacent to $q$: $q$ is connected to two constraints; in class 0, these constraints share one additional variable node, whereas in class 1, they connect to two distinct additional variable nodes. We include several variants of each class so that graph size and construction order alone do not reveal the label. By construction, the task is solvable from the $2$-hop neighborhood of $q$.

\begin{table}[H]
    \centering
    \small
    \setlength{\tabcolsep}{8pt}
    \caption{{Dataset statistics for the synthetic ID-permutation diagnostic.}}
    \label{tab:toy-id-shortcut-stats}
    \begin{tabular}{lcc}
        \toprule
        \textbf{Split} & \textbf{Avg \# Variables / Constraints} & \textbf{Avg Total Nodes} \\
        \midrule
        Train & 25.39 / 21.35 & 46.74 \\
        Test & 25.45 / 21.57 & 47.02 \\
        \bottomrule
    \end{tabular}
\end{table}

We use the same $2$-layer GCN for all methods and compare No-Aug, Uniform, Position, and ColorUID. ColorUID instantiates the Local-UID scheme with $2$-hop local uniqueness. Each graph contributes exactly one supervised example, so accuracy is the fraction of graphs for which the binary label of $q$ is predicted correctly.

To test reliance on arbitrary variable IDs, we apply a counterfactual ID-permutation intervention. The graph structure, labels, and ColorUID colors are kept fixed, while only the variable IDs used by Position are permuted. Thus, the ID-permuted test split contains the same graphs as the ordinary test split and differs only in the identifiers visible to Position. Because the label depends only on the local neighborhood of $q$, degradation under this intervention is evidence of reliance on the ID signal rather than on the task-relevant local structure. ColorUID preserves the local information needed for prediction and is invariant to these global node-order artifacts.

\begin{table}[H]
    \centering
    \small
    \setlength{\tabcolsep}{5pt}
    \caption{{Results for the synthetic ID-permutation diagnostic. Accuracy is reported as mean $\pm$ standard deviation (higher is better).}}
    \label{tab:toy-id-shortcut-results}
    \begin{tabular}{lccc}
        \toprule
        \textbf{Method} & \textbf{Train Acc} & \textbf{Test Acc} & \textbf{ID-Permuted Test Acc} \\
        \midrule
        No-Aug & $80.44{\pm}6.91$ & $80.18{\pm}7.85$ & $80.18{\pm}7.85$ \\
        Uniform & $74.80{\pm}12.77$ & $56.46{\pm}3.53$ & $56.46{\pm}3.53$ \\     
        Position & $89.56{\pm}1.89$ & $89.60{\pm}1.06$ & $51.38{\pm}0.64$ \\
        \midrule
        ColorUID & \best{$97.84{\pm}0.39$} & \best{$97.22{\pm}0.29$} & \best{$97.22{\pm}0.29$} \\
        \bottomrule
    \end{tabular}
\end{table}

Table~\ref{tab:toy-id-shortcut-results} shows that Position performs well on the standard test split, but its accuracy drops to near chance under the ID-permutation intervention. In contrast, ColorUID remains unchanged under the same intervention. This suggests that, in this controlled setting, Position-style Global-UID signals can create shortcut reliance on arbitrary variable IDs, while Local-UID is more stable because it does not depend on global identifiers. The Uniform result is secondary; the key comparison is between Position's degradation under ID permutation and ColorUID's invariance. This experiment is illustrative rather than exhaustive: it does not imply that Local-UID is universally superior to all random-ID or positional-encoding variants across tasks, graph families, or hyperparameter settings. It supports the narrower claim that Position-style Global-UID signals can be brittle in this setting, whereas Local-UID provides a local alternative aligned with the task-relevant neighborhood.

\section{{Comparison with an Alternative Coloring Method: Weak Coloring}}
\label{app:weak-coloring}
{
We compare against the weak coloring scheme of \citet{sato2019approximation}, which is closely related to GNN expressivity for combinatorial problems. A weak 2-coloring is a function $c: V \to \{0,1\}$ such that for every node $v \in V$, there exists a neighbor $u \in N(v)$ with $c(v) \neq c(u)$.

We evaluate this WeakColor augmentation on BPP, a standard ILP benchmark, using the same Top-$m\%$ error metrics as in the main experiments, and compare it against Non-Aug, ColorUID, and ColorGNN.

\begin{table}[H]
    \centering
    \small
    \setlength{\tabcolsep}{3pt}
    \caption{{Weak coloring comparison on the BPP dataset. Results are Top-$m\%$ errors reported as mean $\pm$ standard deviation (lower is better).}}
    \label{tab:weak-coloring-bpp}
    \begin{tabular}{lccccc}
        \toprule
        \textbf{Method} & \textbf{Top-30\%} & \textbf{Top-50\%} & \textbf{Top-70\%} & \textbf{Top-90\%} & \textbf{Top-100\%} \\
        \midrule
        Non-Aug & $5.989{\pm}0.120$ & $9.973{\pm}0.156$ & $14.775{\pm}0.230$ & $18.728{\pm}0.299$ & $28.633{\pm}0.462$ \\
        WeakColor & $6.066{\pm}0.013$ & $10.061{\pm}0.008$ & $14.890{\pm}0.011$ & $18.878{\pm}0.005$ & $28.873{\pm}0.002$ \\
        \midrule
        ColorUID & \best{$0.00{\pm}0.00$} & $0.03{\pm}0.02$ & $0.18{\pm}0.07$ & $2.66{\pm}0.45$ & \best{$12.52{\pm}0.98$} \\
        ColorGNN & \best{$0.00{\pm}0.00$} & \best{$0.00{\pm}0.00$} & \best{$0.09{\pm}0.04$} & \best{$2.40{\pm}0.40$} & $13.13{\pm}1.01$ \\
        \bottomrule
    \end{tabular}
\end{table}

Table~\ref{tab:weak-coloring-bpp} shows that WeakColor has higher mean error than Non-Aug at every reported Top-$m\%$ threshold, whereas ColorUID and ColorGNN outperform both baselines. These results should not be read as evidence that weak coloring is generally ineffective; instead, they suggest a mismatch with this ILP bipartite-graph setting. ILP instances are encoded as bipartite graphs with variable nodes and constraint nodes. On such graphs, a weak 2-coloring can be satisfied by a coarse signal that marks only the two sides of the bipartition. In that case, all variable nodes receive the same weak-color signal, which does not help distinguish variables for solution prediction. Adding such uninformative color features can therefore make learning no easier, and may make it harder, in this setting.

This comparison highlights a distinction between weak coloring and the Local-UID principle used in this paper. Weak coloring only requires each node to have at least one neighbor with the opposite color; it does not require nodes within a GNN receptive field to be distinguishable. By contrast, $d$-hop unique coloring is designed to provide local identity information within each receptive field. This distinction matters for variable-level prediction on ILP bipartite graphs, where separating individual variables is central to the learning task.
}

\section{{Backbone Sensitivity of the Local-UID scheme}}
\label{app:stronger-backbone-check}
\makeatletter
\renewcommand{\fnum@table}{{\tablename~\thetable}}
\makeatother
To test whether the benefit of identifier/color augmentation depends on the backbone used in the main experiments, we run additional experiments on BPP with two stronger architectures: Graphormer~\citep{ying2021transformers} and GAT~\citep{velivckovic2017graph}. Within each backbone group, the data split, optimization setup, epoch budget, and sampling setting are fixed; only the identifier/color injection changes. We compare Non-Aug, Uniform, Position, Orbit, Orbit+, ColorUID, and ColorGNN. 
\begin{table}[t]
    \centering
    \small
    \setlength{\tabcolsep}{3pt}
    \caption{{BPP results with matched Graphormer and GAT backbones. Top-$m\%$ error is reported (lower is better}).}
    \label{tab:stronger-backbone-bpp}
    \resizebox{\linewidth}{!}{
    \begin{tabular}{llccccc}
        \toprule
        \textbf{Backbone} & \textbf{Method} & \textbf{Top-30\%} & \textbf{Top-50\%} & \textbf{Top-70\%} & \textbf{Top-90\%} & \textbf{Top-100\%} \\
        \midrule
        \multirow{7}{*}{Graphormer}
        & Non-Aug & $5.53{\pm}0.71$ & $9.74{\pm}0.68$ & $14.47{\pm}0.12$ & $18.61{\pm}0.09$ & $28.87{\pm}0.00$ \\
        & Uniform & $0.01{\pm}0.01$ & $0.06{\pm}0.02$ & $0.58{\pm}0.14$ & $5.74{\pm}0.38$ & $18.47{\pm}0.26$ \\
        & Position & $0.03{\pm}0.03$ & $0.12{\pm}0.07$ & $0.73{\pm}0.26$ & $5.72{\pm}0.33$ & $18.09{\pm}0.22$ \\
        & Orbit & \best{$0.00{\pm}0.00$} & \best{$0.01{\pm}0.00$} & $0.10{\pm}0.06$ & $3.61{\pm}0.18$ & $14.90{\pm}0.11$ \\
        & Orbit+ & $0.04{\pm}0.06$ & $0.11{\pm}0.18$ & $0.76{\pm}1.25$ & $3.49{\pm}2.24$ & $13.78{\pm}2.14$ \\
        \cmidrule(lr){2-7}
        & ColorUID & \best{$0.00{\pm}0.00$} & $0.01{\pm}0.01$ & \best{$0.06{\pm}0.02$} & \best{$1.99{\pm}0.73$} & $12.35{\pm}1.40$ \\
        & ColorGNN & $0.01{\pm}0.01$ & $0.02{\pm}0.04$ & $0.10{\pm}0.07$ & $2.02{\pm}0.55$ & \best{$11.72{\pm}0.57$} \\
        \midrule
        \multirow{7}{*}{GAT}
        & Non-Aug & $6.38{\pm}0.02$ & $10.46{\pm}0.01$ & $14.59{\pm}0.05$ & $18.79{\pm}0.02$ & $28.87{\pm}0.00$ \\
        & Uniform & $0.24{\pm}0.23$ & $1.21{\pm}1.17$ & $2.93{\pm}2.18$ & $8.98{\pm}1.51$ & $20.36{\pm}3.49$ \\
        & Position & $0.35{\pm}0.15$ & $1.63{\pm}1.23$ & $3.96{\pm}2.80$ & $9.45{\pm}1.37$ & $23.66{\pm}0.13$ \\
        & Orbit & $0.31{\pm}0.18$ & $2.22{\pm}1.18$ & $5.08{\pm}2.39$ & $9.24{\pm}2.48$ & $21.88{\pm}3.37$ \\
        & Orbit+ & $0.13{\pm}0.12$ & $0.47{\pm}0.34$ & $2.31{\pm}1.09$ & $6.71{\pm}1.53$ & $20.29{\pm}4.11$ \\
        \cmidrule(lr){2-7}
        & ColorUID & \best{$0.00{\pm}0.00$} & \best{$0.02{\pm}0.01$} & $0.43{\pm}0.26$ & \best{$5.07{\pm}2.34$} & $18.04{\pm}4.05$ \\
        & ColorGNN & \best{$0.00{\pm}0.00$} & $0.03{\pm}0.02$ & \best{$0.35{\pm}0.21$} & $5.83{\pm}2.13$ & \best{$17.04{\pm}2.77$} \\
        \bottomrule
    \end{tabular}}
\end{table}

Table~\ref{tab:stronger-backbone-bpp} shows that identifier/color-based augmentation remains beneficial with both stronger backbones. With Graphormer, ColorUID achieves the best Top-70\% and Top-90\% errors, while ColorGNN achieves the best Top-100\% error. With GAT, ColorUID and ColorGNN outperform Non-Aug and improve over Uniform, Position, Orbit, and Orbit+ on most thresholds. Graphormer gives the strongest overall results in this comparison, but at a higher computational cost.

Overall, these experiments indicate that the Local-UID/color gains are not an artifact of the original backbone, although they do not establish that gains increase with backbone strength or characterize all expressive GNN architectures.

\section{{Training and Test Losses Across Datasets}}
\label{app:train-test-losses}
We report the training and test losses for the methods evaluated in the paper. All values are computed over four random seeds and reported as mean $\pm$ standard deviation. Because the loss scale is dataset-dependent, comparisons should be made within each dataset rather than across datasets.

\begin{table}[t]
    \centering
    \footnotesize
    \setlength{\tabcolsep}{6pt}
    \renewcommand{\arraystretch}{0.96}
    \caption{{Training and test losses across all reported datasets. Results are grouped by dataset and reported as mean $\pm$ standard deviation over four seeds (lower is better).}}
    \label{tab:train-test-loss-all}
    \begin{tabular}{llcc}
        \toprule
        \textbf{Dataset} & \textbf{Method} & \textbf{Train Loss} & \textbf{Test Loss} \\
        \midrule
        \multirow{20}{*}{BPP}
        & Non-Aug & $93.16{\pm}0.00$ & $93.16{\pm}0.00$ \\
        & Position & $58.21{\pm}8.23$ & $58.26{\pm}8.20$ \\
        & Uniform & $64.41{\pm}5.44$ & $64.55{\pm}6.37$ \\
        & Orbit & $36.20{\pm}0.70$ & $35.48{\pm}0.61$ \\
        & Orbit+ & $35.14{\pm}0.24$ & $34.27{\pm}0.23$ \\
        & $\colorlui$ ($d=1$) & $93.16{\pm}0.00$ & $93.16{\pm}0.00$ \\
        & $\colorlui$ ($d=2$) & $32.90{\pm}3.07$ & $32.65{\pm}2.50$ \\
        & $\colorlui$ ($d=3$) & $44.06{\pm}5.37$ & $43.95{\pm}5.77$ \\
        & $\colorlui$ ($d=4$) & \best{$27.64{\pm}1.16$} & \best{$28.74{\pm}0.69$} \\
        & $\colorlui$ ($d=5$) & $28.74{\pm}0.32$ & $29.59{\pm}0.13$ \\
        & $\colorlui$ ($d=6$) & $28.95{\pm}0.14$ & $29.65{\pm}0.28$ \\
        & $\colorgnn$ & $32.05{\pm}2.86$ & $32.60{\pm}2.09$ \\
        & $\colorgnn$,Orbit & $43.18{\pm}7.89$ & $44.07{\pm}6.89$ \\
        & $\colorgnn$,Orbit+ & $36.19{\pm}2.32$ & $36.21{\pm}2.75$ \\
        & $\colorgnn(\merge)$ & $41.44{\pm}6.24$ & $42.11{\pm}5.65$ \\
        & $\colorgnn(\merge)$,Orbit & $46.87{\pm}0.85$ & $47.54{\pm}0.61$ \\
        & $\colorgnn(\merge)$,Orbit+ & $38.40{\pm}2.03$ & $39.84{\pm}2.19$ \\
        & $\colorgnn(\aggr)$ & $43.18{\pm}2.34$ & $42.45{\pm}1.76$ \\
        & $\colorgnn(\aggr)$,Orbit & $33.91{\pm}2.22$ & $33.66{\pm}1.94$ \\
        & $\colorgnn(\aggr)$,Orbit+ & $32.39{\pm}0.79$ & $31.71{\pm}0.57$ \\
        \midrule
        \multirow{7}{*}{BIP}
        & Non-Aug & $341.32{\pm}0.00$ & $341.32{\pm}0.00$ \\
        & Position & $297.59{\pm}2.29$ & $297.51{\pm}1.86$ \\
        & Uniform & $307.62{\pm}5.04$ & $304.98{\pm}4.07$ \\
        & Orbit & $279.80{\pm}1.45$ & $278.26{\pm}0.70$ \\
        & Orbit+ & $277.24{\pm}0.52$ & $276.96{\pm}0.36$ \\
        & $\colorlui$ & $265.43{\pm}0.49$ & $266.21{\pm}0.43$ \\
        & $\colorgnn$ & \best{$258.00{\pm}2.05$} & \best{$261.39{\pm}1.83$} \\
        \midrule
        \multirow{7}{*}{SMSP}
        & Non-Aug & $906.86{\pm}2.26$ & $906.93{\pm}3.36$ \\
        & Position & $615.92{\pm}11.75$ & $615.61{\pm}1.28$ \\
        & Uniform & $647.67{\pm}40.51$ & $649.37{\pm}38.87$ \\
        & Orbit & $571.63{\pm}7.89$ & $572.33{\pm}6.25$ \\
        & Orbit+ & $553.39{\pm}4.02$ & $553.72{\pm}4.01$ \\
        & $\colorlui$ & $392.21{\pm}7.06$ & $402.72{\pm}12.40$ \\
        & $\colorgnn$ & \best{$388.84{\pm}13.31$} & \best{$401.63{\pm}12.79$} \\
        \midrule
        \multirow{5}{*}{IS}
        & Non-Aug & $564.80{\pm}0.26$ & $564.65{\pm}0.39$ \\
        & Position & $390.53{\pm}2.00$ & $387.20{\pm}1.22$ \\
        & Uniform & $391.83{\pm}2.26$ & $389.84{\pm}1.23$ \\
        & $\colorlui$ & $162.07{\pm}0.87$ & $162.38{\pm}1.10$ \\
        & $\colorgnn$ & \best{$161.48{\pm}0.68$} & \best{$161.87{\pm}1.13$} \\
        \midrule
        \multirow{5}{*}{CA}
        & Non-Aug & $740.48{\pm}0.21$ & $740.89{\pm}0.31$ \\
        & Position & $673.09{\pm}0.88$ & $674.60{\pm}1.04$ \\
        & Uniform & $674.60{\pm}1.14$ & $678.90{\pm}0.69$ \\
        & $\colorlui$ & $657.33{\pm}0.74$ & $662.89{\pm}1.82$ \\
        & $\colorgnn$ & \best{$649.29{\pm}0.65$} & \best{$651.75{\pm}1.14$} \\
        \bottomrule
    \end{tabular}
\end{table}

The reported losses are consistent with the main evaluation trends. On each dataset, the best local/color-based configuration reduces both training and test losses relative to Non-Aug and to the global Position/Uniform baselines. On BPP, the lowest losses are obtained by $\colorlui$ with $d=4$; on BIP, SMSP, IS, and CA, they are obtained by $\colorgnn$. The training and test losses are close for most methods, indicating that the observed gains are not simply lower training losses without corresponding improvements on the test split.

\section{{Average Coloring Time Across Datasets}}
\label{app:coloring-time}
{
Table~\ref{tab:avg-coloring-time} reports the average number of nodes and the average coloring time for each dataset. The timings use the greedy 1-hop unique coloring preprocessing used in the experiments and are averaged over the corresponding benchmark instances.

\begin{table}[H]
    \centering
    \small
    \setlength{\tabcolsep}{8pt}
    \renewcommand{\arraystretch}{1.05}
    \caption{{Average graph size and coloring time across datasets.}}
    \label{tab:avg-coloring-time}
    \begin{tabular}{lccccc}
        \toprule
        \textbf{Dataset} & \textbf{BPP} & \textbf{BIP} & \textbf{SMSP} & \textbf{IS} & \textbf{CA} \\
        \midrule
        \textbf{Avg. \# nodes} & 460 & 1278 & 45854.1 & 2100.00 & 7896.00 \\
        \textbf{Coloring time (s)} & 0.002 & 0.028 & 2.952 & 0.003 & 0.043 \\
        \bottomrule
    \end{tabular}
\end{table}
}

\end{document}